# Advanced Fuzzy Cognitive Maps: State Space and Rule-Based methodology for Coronary Artery Disease detection

**Ioannis D. Apostolopoulos [1*], Peter P. Groumpos [2], Dimitris I. Apostolopoulos**

[1] University of Patras, Medical School, Department of Medical Physics, Rio, Achaia, PC 26504, Greece
ece7216@upnet.gr
[2] University of Patras, Department Electrical and Computer Engineering, Rio, Achaia, PC 26504, Greece;
groumpos@ece.upatras.gr
[3] University of Patras, Medical School, Department of Nuclear Medicine, Rio, Achaia, PC 26504, Greece. Tel.: +30 6972122372, email: dimap@med.upatras.gr

**\*** Correspondence: Ioannis D. Apostolopoulos - ece7216@upnet.gr

**Abstract**
In this study, the recently emerged advances in Fuzzy Cognitive Maps (FCM) are investigated and employed, for achieving automatic and non-invasive diagnosis of Coronary Artery Disease (CAD). A Computer-Aided Diagnostic model for a precise prediction of CAD using State Space Advanced FCM (AFCM) approach is proposed. Also, a rule-based mechanism is incorporated, to further increase the knowledge of the system and the interpretability of the decision mechanism. The proposed method is tested utilizing a CAD dataset from the Laboratory of Nuclear Medicine of the University of Patras. More specifically, two architectures of AFCMs are designed, and parameter testing is performed. Furthermore, the proposed AFCMs, which are based on novel equations proposed in literature, are compared with the traditional FCM approach. The experiments highlight the effectiveness of the AFCM approach, which obtained an accuracy of 78.21%, achieving an increase of seven percent (+7%) on the classification task, and obtaining 85.47% accuracy. It is demonstrated that the AFCM approach in developing Fuzzy Cognitive Maps outperforms the conventional approach, while it constitutes a reliable method for the diagnosis of Coronary Artery Disease.

**Highlights**

- Advanced Fuzzy Cognitive Maps approach (State Space) is proposed for automatic diagnosis of Coronary Artery Disease
- Rule-Based approach is embedded to the framework to increase the performance
- 

## 1. Introduction

Medical Decision-Making problems are complex. Traditionally, physicians are requested to handle information of different nature, e.g. patient's history, clinical diagnostic tests, medical images, and demographic characteristics. The interpretation of the results involves uncertainty, which plays a critical role for the decision-making to a wide and diverse set of fields. Approaching dynamic systems that involve uncertainty is demanding, therefore engineers and decision-makers face difficulties (**Bhattacharyya 2020**).

Clinical Decision Support Systems (CDSS) have been proposed the past years, to overcome those issues. A CDSS is an expert system that provides decision support and serves as a diagnostic and treatment reference for clinicians or patients. CDSSs incorporate any available patient related clinical data and the experts' knowledge to perform a reasoning analysis.

A lot discussion is held regarding the decision mechanisms relating the inputs and the desired outputs of the systems. Fuzzy Cognitive Maps (FCMs) constitute a simple computational and graphical methodology to represent complex problems. FCMs' decision-making mechanism is a unique method of handling the parameters of a desired decision.

FCMs were introduced by Kosko in 1986 in order to represent the causal relationship between concepts and analyze inference patterns (**Kosko 1986, Kosko 1998**). They take advantage of the knowledge and the experience of experts, offering them an alternative way of addressing the problems, yet in the same way a human mind does. This is achieved by using a conceptual procedure, which can include ambiguous of fuzzy descriptions (**Bourgani 2014**). Recently, Fuzzy Cognitive Maps were employed to describe and solve medical problems (**Papageorgiou 2008a, Papageorgiou 2011; Giabbanelli 2012**).

State Space Advanced Fuzzy Cognitive Maps (AFCM) are an evolution of the classic methodology, which promises more precise results for a large variety of complex systems. This methodology, which is thoroughly analyzed in (**Mpelogianni 2018a**), is presented in the following subsections, and it overcomes some issues of the classic FCMs. Those issues are: (a) the presence of concepts of different nature was ignored in the traditional approach and, (b) the utilization of the classic sigmoid normalization function was fuzzing the system, especially in cases of the existence of several concepts, due to the fact that the output value was always transformed to a number near one. Despite the fact that the novel strategy in (**Mpelogianni 2018**) is intended for applications where time iteration steps play the most vital role in the behavior of the system, they can also be employed to applications where time does not play any role.

As to our best knowledge, AFCM have not been applied and evaluated to medical diagnosis. In this work, we employ the newly emerged State Space FCMs for the automatic and non-invasive prediction of Coronary Artery Disease (CAD). The Coronary Artery Disease is caused when the atherosclerotic plaques load, namely fill, in the lumen of the blood vessels of the heart, which are named coronary arteries, and they obstruct the blood flow to the heart. According to the World Health Organization, 50% of deaths in European Union is caused by Cardiovascular Diseases (CVD), while 80% of the premature heart diseases and strokes can be prevented (**WHO 2014**). Diagnosing CAD in a non-invasive way is an open challenge, despite the massive number of researches been made so far (**Dilsizian 2014; Soni 2011, Apostolopoulos 2020**). Several research works are focused in proposing Rule-Based Machine Learning or Fuzzy Logic utilizing publically available datasets (**Abdar 2019, Ghiasi 2020, Setiawan 2020, Hossain 2020**). The present work is not focused on performing a thorough analysis of the best-performing approaches. Besides, comparisons are not meaningful because different sets of data of various sizes are utilized. We focused our work in improving and confirming the effectiveness of the AFCM approach over the classic, which was implemented in a recent work by the authors (**Apostolopoulos 2020**).

Designing AFCM models requires the collaboration between subject-matter experts and modelers. The experts suggest and aid for the design of the AFCM, for it to be user-friendly, scientifically acceptable, and interpretable. In this work, the Nuclear Medicine staff suggested that the causal relationships between concepts in (**Apostolopoulos 2020**) were unable to explain some unique and complex connections. Hence, the proposed models were designed to function with the aid of some universal rules. The Rule-Embedded AFCM (RE-AFCM), which is proposed in this work, is evaluated on a real patient-candidate database from the Laboratory of Nuclear Medicine of the University of Patras. Aiming to improve and evaluate AFCM's performance, we experiment with the new and the traditional equations, as well as with different activation functions and architectures. The optimal parameters are defined through the The results demonstrate the effectiveness of the newly emerged State Space approach for the development of FCMs, improving the accuracy by 7%. It is also demonstrated that RE-AFCM can be applied to medical problems, with the appropriate modifications and with respect to the nature of the inputs.

## 2. Methods

2.1 Traditional FCM theory

FCMs (**Kosko 1986**) constitute a computational methodology able to examine situations which the human thinking process involves fuzzy or uncertain descriptions. FCMs consist of a graphical representation through a signed directed graph, which includes feedback consisting of nodes and weighted arcs (**Kosko 1986, Groumpos 2000, Felix 2019**). Each node in the graph represents a concept used to describe the cause and effect relationship. The nodes are connected by weighted arcs representing the actual interconnections. Each concept Ci (i.e., the variable of the system) is characterized by a number representing its values, and is calculated through the transformation of a fuzzy value, or the fitting of a numeric value, to the desired interval [0,1]. The initial weight values are defined by the experts of the domain and, therefore, they are linguistic variables, or rule-type statements. Through a defuzzification procedure the linguistic variables are transformed into numeric weights. In this way, FCMs embody the accumulated knowledge and experience from experts (**Groumpos 2018**). An example is given in Figure 1.

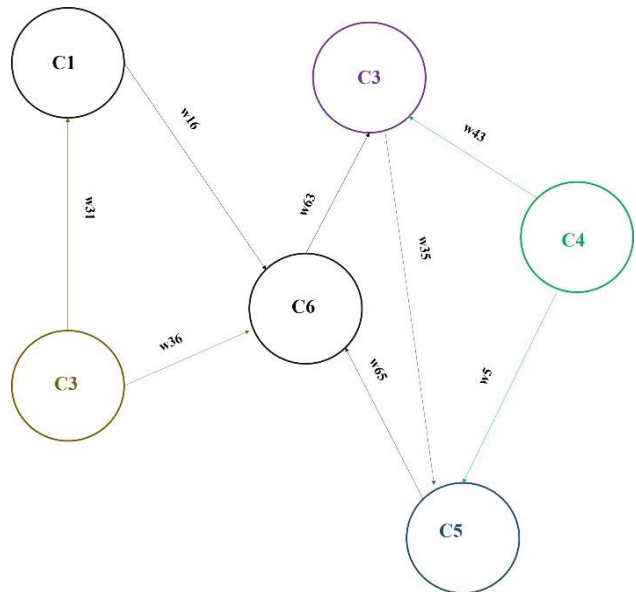

**Fig.1** Fuzzy Cognitive Map.

The degree of influence between the two concepts is indicated by the absolute value of $W_{ij}$. During the simulation, the value of each concept ($C_i$) is calculated as shown in (**Salmeron 2019, Papageorgiou 2008b**).

2.2 State Approach

In the classic FCM representation, all concepts and parameters are treated and calculated regardless of their different nature. However, even when a system is described in fuzzy way, the main concept is the same. In the classic control theory, which shall be employed to address the matter, a separation of the concepts is suggested (**Ogata 1970; Mpelogianni 2018b**), as follows:

- **Input Concepts:** The inputs of the system, (u)
- **State Concepts:** The concepts describing the operation of the system, (x)
- **Output Concepts:** The concepts describing the outputs of the system, (y).

A simple representation of the system is achieved by a block diagram, which is presented in (**Mpelogianni 2018b**), and depicted in Figure 2.

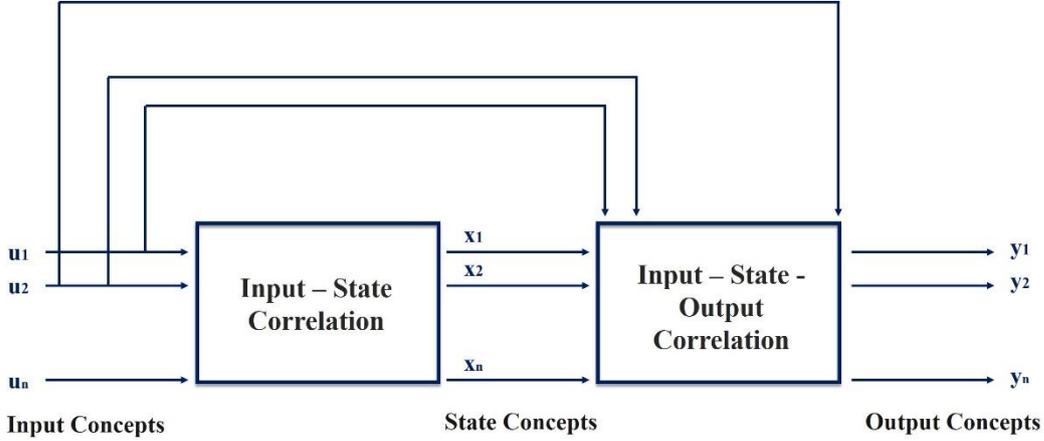

**Fig.2** Block Diagram of the separated concepts.

In this way, more accurate knowledge of the system is obtained, because the state concepts are separated from the input concepts. Moreover, introducing state concepts aids in a more complete characterization and modeling of human decision-making process. The proposed separation facilitates the understanding of the system's operation, and the simpler calculation of the concepts' values.

*2.2.1 Mathematical Expressions of the State Space Advanced Approach*

Separating the concepts into categories enables the calculation of their values in a more distributed way. The initial weight matrix, which describes all relations, shall be divided into smaller ones in order to correspond them to each concept category. The equations (eq. 1, eq. 2)

$$X[k+1] = Ax[k] + Bu[k] \quad (1)$$

$$Y[k+1] = Cx[k] + Du[k] \quad (2)$$

Shall now be used to calculate the variation, caused by the change in the input and state concepts, to the state and output concepts, at each time step [k]. In equations (1), and (2), A, B, C and D are individual weight matrixes derived from the initial; The elements of matrix A depend on the states' weights, while the elements of matrix B show how each input concept affects the state concepts of the system. Matrix C embodies the relation between the states and the outputs, while matrix D incorporates the direct affection of input concepts to output concepts (**Ogata 1970**).

Since equations (1) and (2) are used to compute the variation caused between the concepts, it is more accurate to express them as follows:

$$\Delta X[k+1] = Ax[k] + Bu[k] \quad (3)$$

$$\Delta Y[k+1] = Cx[k] + Du[k] \quad (4)$$

Using equations (3) and (4), the expression of the actual values of X[k+1] and Y[k+1] is in equations (5) and (6):

$$X[k+1] = X[K] + \frac{\Delta X[\kappa+1]}{\sum_{j=1, j\neq i}^{n} |Wji|} \quad (5)$$

$$Y[k+1] = Y[K] + \frac{\Delta Y[\kappa+1]}{\sum_{j=1, j\neq i}^{n} |Wji|} \quad (6)$$

One more advantage of the mentioned technique is the transparency and decomposability of the system.

*2.2.2 Activation and Normalization Functions*

In order to apply the AFCM methodology, the values of all the input concepts must lay between the interval [0,1], where 0 denotes that the value of the concept is very small and 1 that the value is very big. There are many proposed activation functions to perform the aforementioned task. In this work we utilize the newly proposed alteration to the classic sigmoid function, as introduced in (**Mpelogianni 2016**). We use this function for the inner calculations (i.e. the inputs and the states). The function is explained below:

$$f(x) = m + \frac{M-m}{1+e^{-r*(x-to)}} \quad (7)$$

In equation (7), m is the lower limit of the curve, M is the upper limit of the curve, r is the slope of the curve and $t_o$ is the symmetry to the y axis. We refer to equation (7) as SigmoidN.

For the final classification between "Healthy" or "Diseased", we also use a Softmax classifier (**Jang 2016**), to evaluate and compare the different methodologies.

*2.2.3 Fuzzification and de-fuzzification process*

An overview of the fuzzification and the de-fuzzification process shall be presented.
- Step 1: The experts assign the weight values (verbal). Those weight values shall be de-fuzzified into crisp numbers in space [0,1].
- Step 2: The concepts initial values may be verbal or numerical. As in step 1, the verbal inputs shall be de-fuzzified into crisp numbers in space [0,1].
- Step 3: When reaching a stable state after a specific amount of iterations, the model's output values are numerical in the desired space. Those values shall then be fuzzified and substituted with verbal ones (e.g. 0.85 equals to "Definitely Abnormal Situation") The process of fuzzification and de-fuzzification shall be done with different methods, such as the Center of Area (COA) (**Axelrod 2015**), or Center of Gravity (COG) (**Jang 1997**). For the normalization of the output values, both sigmoid or tanh functions may be utilized.

2.3. The proposed models

*2.3.1. The overall procedure*

Two Nuclear Medicine doctors were pooled to define the concepts, the interconnections and the outputs of the system. The system was designed to meet the needs of the Department of Nuclear Medicine of Patras. Therefore, the input concepts of the system reflect the department's approach in diagnosing CAD. The relationships between the concepts of the system are described by the doctors in a linguistic way, thus allowing freedom to explain the interconnections the way experts prefer. The rest of the procedure, i.e., the de-fuzzification of the inputs and weights, is performed by the engineers. We can summarize the steps of the process as follows: (a) the experts decide the inputs of the system and their possible values, (b) each expert assigns a specific verbal weight between the concepts he/she believes that share a connection, (c) the doctors specify the specific rules describing the system and provide the necessary documentation that confirms the rules, (d) the doctors in collaboration with the engineers define the state concepts of the system, (e) the input values, the state and the weight values as well as the rules are transformed from nominal to numeric or binary form, following specific mathematical expressions, and (f) the numeric value of the output is calculated. The normalization function returns the score, or the probability of the subject.

Based on the aforementioned approach in designing FCMs, the concepts of the AFCM shall be divided into inputs, states and outputs.

*2.3.2 Concept Definition*

The global input concepts, i.e., before the specific input and state concepts are defined, as well as their possible values as defined by the experts are given in the Table 1.

**Table 1.** Initial concepts of the proposed AFCM

| | Attributes | Values |
|---|---|---|
| A1 | typical angina pectoris | yes, no |
| A2 | atypical angina pectoris | yes, no |
| A3 | atypical thoracic pain | yes, no |
| A4 | dyspnea on exertion | yes, no |
| A5 | asymptomatic | yes, no |
| A6 | gender - male | yes, no |
| A7 | gender - female | yes, no |
| A8 | age <40 | yes, no |
| A9 | age [40-50] | yes, no |
| A10 | age [50-60] | yes, no |
| A11 | age >60 | yes, no |
| A12 | known CAD | yes, no |
| A13 | previous stroke | yes, no |
| A14 | peripheral arterial disease | yes, no |
| A15 | smoking | yes, occasionally, no |
| A16 | arterial hypertension | yes, no |
| A17 | dyslipidemia | yes, no |
| A18 | obesity | yes, relatively, no |
| A19 | family history | yes, no |
| A20 | diabetes | yes, no |
| A21 | chronic kidney failure | yes, no |
| A22 | electrocardiogram normal | yes, no |
| A23 | electrocardiogram abnormal | yes, no |
| A24 | echocardiogram normal - doubtful | yes, no |
| A25 | echocardiogram abnormal | little, abnormal, definitely abnormal |
| A26 | treadmill exercise test normal | yes, no |
| A27 | treadmill exercise test abnormal | abnormal, definitely abnormal |
| A28 | dynamic echocardiogram normal | yes, no |
| A29 | dynamic echocardiogram abnormal | doubtful, abnormal, definitely abnormal |
| A30 | scintigraphy normal - doubtful | yes, no |
| A31 | scintigraphy abnormal | Little abnormal , medium abnormal, abnormal, definitely abnormal |

The state concepts were discussed and designed in collaboration with the experts. The state concepts may be constituted by the following expressions:

- **A32:** Predisposing factors
- **A33:** Recurrent Diseases
- **A34:** Demographic Characteristics
- **A35:** Diagnostic Tests

The reader should note that we change the number of state concepts for the experiments, to inspect the effectiveness of each state-concept. For example, we design a AFCM with only one state concept before we proceed to adding more state concepts. It is also mentioned that when the system is designed with state concepts, some of the input concepts are directly connected to the state concepts and not to the output.

The proposed system classifies the instances to "Healthy" or "Diseased"; thus, the system shall have two possible classes as outputs. We propose two approaches for the final classification. The first approach suggests a single output concept (Out), the value of which describes the probability of infection. For this approach the SigmoidN is utilized through the entire process, due to the fact that it yields better results, as shown in Results section. The second approach suggests two output concepts to be inserted, referred to as "out_healthy", and "out_diseased". They present each class's score. A softmax classifier is utilized to classify each instance, based on the score of each class. For the activation of the concepts, we experiment with both the classic sigmoid function and with the signoidN, with parameters set at M=1, m=-1, $t_o = 0$, r=1.

*2.3.3 Interconnections*

The interconnection weights between nodes shall be undertaken by experts, in cooperation with each other. The possible linguistic values of the weights shall be: Very Weak (VW), Weak (W), Medium (M), Strong (S), Very Strong (VS). The weights may also take negative values, e.g. "-VS".

These values are then defuzzied and a corresponding numerical value will be assigned to each one (**Runkler 1996**). We provide corresponding tables for equations 3 and 4:

Equation 3, Matrix A (input - input): Table 2
Equation 3, Matrix B (input - state): Table 3
Equation 4, Matrix C (output - input): Table 4
Equation 4, Matrix D (state - output): Table 5

**Table 2.** Matrix A of the equation 3. Due to size constrains, in this table, only the inputs that have connections with each other are depicted. The original size of Matrix A is 31x31, with zero diagonal values.

| Concept | Concepts affected | Weights (effect of concept A6 on A22-A31) |
|---------|-------------------|-------------------------------------------|
| A7 | A22 – A31 | +W, -W, +VW, -VW, +W, -W, +W, -W, +W, -W, |

The AFCM's table A embodies the interconnection between the input concepts. In the proposed system, there are relations between the input concept A7 (female) and the concepts describing the diagnostic tests (A22 – A31). Those relations exist for every AFCM designed in the experiments.

Table 3 presents the interconnections between the inputs and the states of the system. That is, which inputs may have an influence on the states of the model. In our case, the concepts A12, A13, A15, and A16-A19 are defining the state A32 (predisposing factors). The concepts A14, A20 and A21 define the state A33 (recurrent diseases). Concepts A6-A11 are constituting the state A34 (Demographic Characteristics). Finally, concepts A22-A31 are constituting the state A35 (Diagnostic Tests).

**Table 3.** Contents of Matrix B of equation 3. The original Matrix B is 4x31, where four represents the four state concepts and thirty-one refers to every input concept. When an input concept is not related to one of the state concepts, each value in Matrix B is zero.

| State | Includes concepts | Weights (effect of input concepts on state concept) |
|-------|-------------------|-----------------------------------------------------|
| **A32:** Predisposing factors | A12, A13, A15, A16-A19 | M, M, W, M, VW, W, VW |
| **A33:** Recurrent Diseases | A14, A20, A21 | M, M, W |
| **A34:** Demographic Characteristics | A6 – A11 | M, -S, -VS, -W, W, S |
| **A35:** Diagnostic Tests | A22 – A32 | -M, M, -W, M, -S, W, -W, M, -VS, S |

In Table 4, the direct relation between the inputs and the outputs is presented. Please note that in this case, the inputs defining the state concepts do not directly affect the output(s).

**Table 4.** Preview of Contents of Matrix C of equation 3. Due to constraints of size, only 5 examples are given. The original Matrix C is 13x1 for one output concept and 13x2 in case of two output concepts. The size of this matrix is 13, due to the fact that only relations between input concepts and the output are expressed. Realtions between input concepts and states are expressed by Table 3.

| Input | Case: Single Output | Case: Two classes | |
|---|---|---|---|
| | | out_healthy | out_diseased |
| A1 | VS | 0 | VS |
| A2 | M | 0 | M |
| A3 | W | 0 | W |
| A4 | W | 0 | W |
| Aj | -S | S | 0 |

In Table 5, the system's Matrix D is presented. Matrix D, contains the connection of the state concepts with the output concepts. In the particular case, interconnections between the states do not exist.

**Table 5.** Contents of Matrix D of equation 3. For one output concept the original matrix is 4x1, while for two output concepts the size is 4x2.

| State | Single Output | Two classes | |
|---|---|---|---|
| | | out_healthy | out_diseased |
| A32 | S | If negative then S, else 0 | If positive then S, else 0 |
| A33 | VS | If negative then VS, else 0 | If positive then VS, else 0 |
| A34 | S | If negative then S, else 0 | If positive then S, else 0 |
| A35 | VS | If negative then VS, else 0 | If positive then VS, else 0 |

All Matrixes A,B,C and D are utilized to calculate the final values of the equations 3 and 4 to define the values of ΔX [k + 1] and ΔY [k + 1]. Those values define the final values of X [k + 1] and Y [k + 1], based on equations 5 and 6.

*2.3.4 Rules*

As mentioned above, modern decision-making problems involve complex relationships, and often a relation between concepts cannot only be explained with the casual FCM strategy (i.e., assigning a specific weight). In this study, a RE-AFCM model is designed, allowing the embedding of more complex expressions (rules).

To achieve this, the addition of a hidden mechanism, referred to as "rule-activator" is proposed. The experts suggested a set of rules to be applied to the model, in order to increase its knowledge. Their suggestions were derived from either recent published guidelines (**Levine 2016**), or their experience. Those rules are presented in Table 6. Some of those rules involve deactivation of specific concepts. Negating or deactivation a concept involves removing its value and the corresponding weight from the system.

**Table 6.** Rules and their effect on systems concepts, inputs and outputs

| Rule | Concepts affected |
|---|---|
| **If Definitely Abnormal Scintigraphy** | The weight of Scintigraphy is significantly increased (+50%) before the iterations |
| **If ECG is Normal and Scintigraphy is Normal** | The weights of both tests are relatively increased (+20%) before the iterations |
| **If previous Stroke** | Negate the Gender Discrimination and de-activate the attribute from the system, before the first iteration step |
| **If Known CAD** | Negate the family history affection before first iteration step |
| **If absense of Diabetes, Known CAD and Previous Stroke** | Increase the weights of the diagnostic test, if the result is "Normal" (+20%) |
| **If the subject is asymptomatic with abnormal Scintigraphy and at least one more abnormal diagnostic test** | Decrease the weight of concept A5 (asymptomatic) by 25% |

## 3. Results

*3.1. The dataset of the study*

The database of the particular study consisted of 303 patient cases, all recorded at the Department of Nuclear Medicine of Patras, in Greece. Most of instances were recorded during the last 8 years. All the patient cases had been pointed to Surgical Coronary Angiography. For every instance, therefore, a final medical report confirming or denying the presence of the disease is available. For the characterization of the instances, the stenosis of the coronary artery is the only criterion, which is obtained by the above mentioned invasive diagnostic test. Patients with stenosis equal or above 70% were labeled as diseased, whereas the rest were labeled as healthy.

The dataset contains 116 healthy cases and 187 diseased cases. Male instances are 266 and Female instances are 37. The attributes of the dataset used for the experiments are corresponding to the factors influencing the diagnosis of CAD. The dataset contains every possible diagnostic test a patient may undergo at the clinical section of the experts' laboratory.

The medical reports regarding the diagnostic test were translated in numeric values with the appropriate staging. This approach was supervised by medical staff. The attributes regarding the patient's history and condition (i.e. smoking) were also in need of preprocessing in order to express the linguistic values (i.e. "smoker") into binary format.

*3.2. Evaluation Criteria and results*

We extensively evaluate our system, therefore, more criteria besides accuracy will be employed. The evaluation criteria of the system shall be the following: (a) Accuracy based on the whole dataset, (b) True Positives, (c) False Positives, (d) False Negatives, (e) True Negatives, (f) Sensitivity, and (g) Specificity.

*3.3. Experiment setups*

For the experiments, we propose several cases. In each case, we employ modified architectures and parameters. A summary of the experiment setups is given below:

- **Case 1:** Traditional FCM, with single output, and sigmoid activation function.
- **Case 2:** Traditional FCM, with two output classes, sigmoid activation function, and Softmax Classifier.
- **Case 3:** AFCM, with single output, two states (A32, A33), sigmoid activation function, no rule activator concept.
- **Case 4:** AFCM, with single output, two states (A32, A33), SigmoidN activation function, no rule activator concept.
- **Case 5:** AFCM, with two output classes, two states (A32, A33), SigmoidN activation function, no rule activator concept, and Softmax Classifier.
- **Case 6:** AFCM, with single output, two states (A32, A33), tanh activation function, no rule activator concept.
- **Case 7:** RE-AFCM, with single output, two states (A32, A33), SigmoidN activation function, no rule activator concept.
- **Case 8:** RE-AFCM, with two output classes, two states (A32, A33), SigmoidN activation function, no rule activator concept, and Softmax Classifier.
- **Case 9:** RE-AFCM, with single output, three states (A32, A33, A34), SigmoidN activation function, with rule activator concept.
- **Case 10:** RE-AFCM, with single output, four states (A32, A33, A34, A35), SigmoidN activation function, rule activator concept.

For cases 1 and 2, we use the FCM proposed in (Apostolopoulos 2017). Table 7 presents the results for the different cases.

Table 7. Accuracy, Sensitivity, and Specificity for the different experiment setups

| Case | Accuracy | Sensitivity | Specificity |
|---|---|---|---|
| Case 1 | 77.34% | 81.55% | 68.02% |
| Case 2 | 78.21% | 83.96% | 68.97% |
| Case 3 | 79.20% | 86.63% | 67.24% |
| Case 4 | 80.19% | 87.70% | 68.10% |
| Case 5 | 79.86% | 87.16% | 68.10% |
| Case 6 | 70.95% | 82.8% | 51.7% |
| Case 7 | 81.84% | 83.42% | 79.31% |
| Case 8 | 80.19% | 87.70% | 68.10% |
| **Case 9** | **85.47%** | **89.3%** | **79.31%** |
| Case 10 | 80.85% | 83.42% | 76.72% |

The reader shall notice that Case 9 does not include the diagnostic tests as state-concept. In this case, each diagnostic test is considered as input concept, which directly affects the output. The same procedure takes place in Cases 3 to 8.

The results highlight the effectiveness of the new equations for computing the concepts' values, as it is demonstrated by the increase in accuracy for Case 3 to Case 10. Moreover, the results clarify that the new proposed activation function, SigmoidN, is more preferable for this task over the sigmoid. Also, the combination of single output with SigmoidN, results in an improvement of performance, compared to the combination of two class outputs, and Softmax Classifier. Finally, the addition of rules further improved the knowledge of the system.

Based on the results, we conclude that the proposed RE-AFCM outperforms the traditional FCM approach. What is more, for the specific task, the optimal parameters shall be an architecture consisting of three states, a single output, and the SigmoidN function. Table 9 presents the confusion matrix of Case 9 (the optimal case).

Table 9. Confusion Matrix of Case 9.

|  | Disease d(D+) | Healthy (D-) | Total |
|---|---|---|---|
| **Predicted Diseased** | 167 | 24 | 191 |
| **Predicted Healthy** | 20 | 92 | 112 |
| **Total** | 187 | 116 | 303 |

The confusion matrix corresponds to sensitivity of 89.3%, specificity of 79.3%, PPV of 87.43% and NPV of 82.14%. The overall accuracy is 85.47%. An overview of the RE-AFCM (case 9) is depicted in Figure 3.

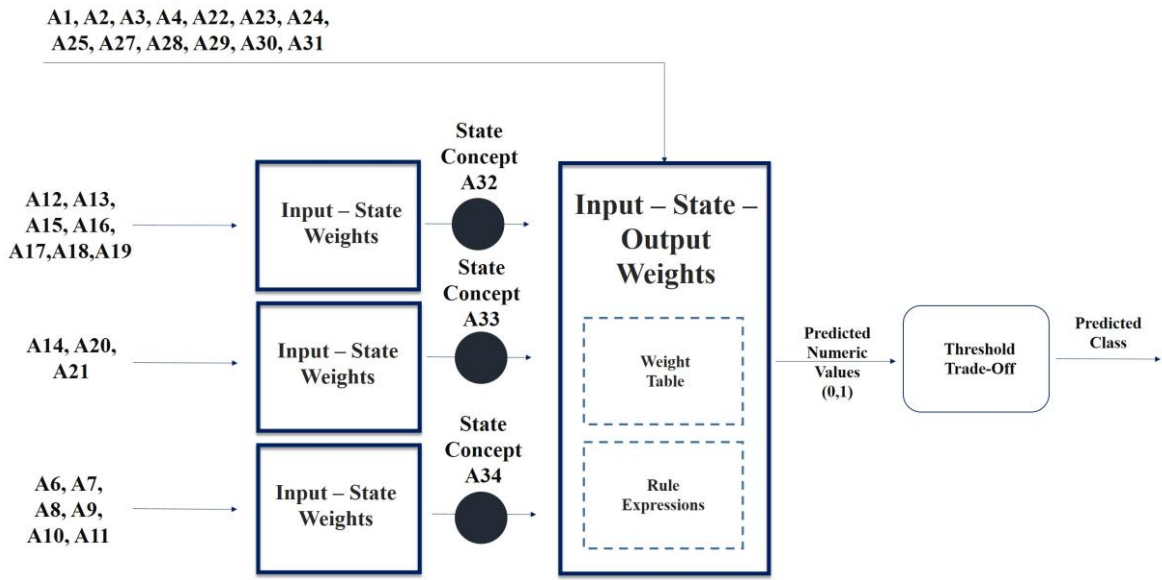

**Fig.3** RE-AFCM (Case 9) overview

The reader can observe the final form of the proposed AFCM in Figure 4.

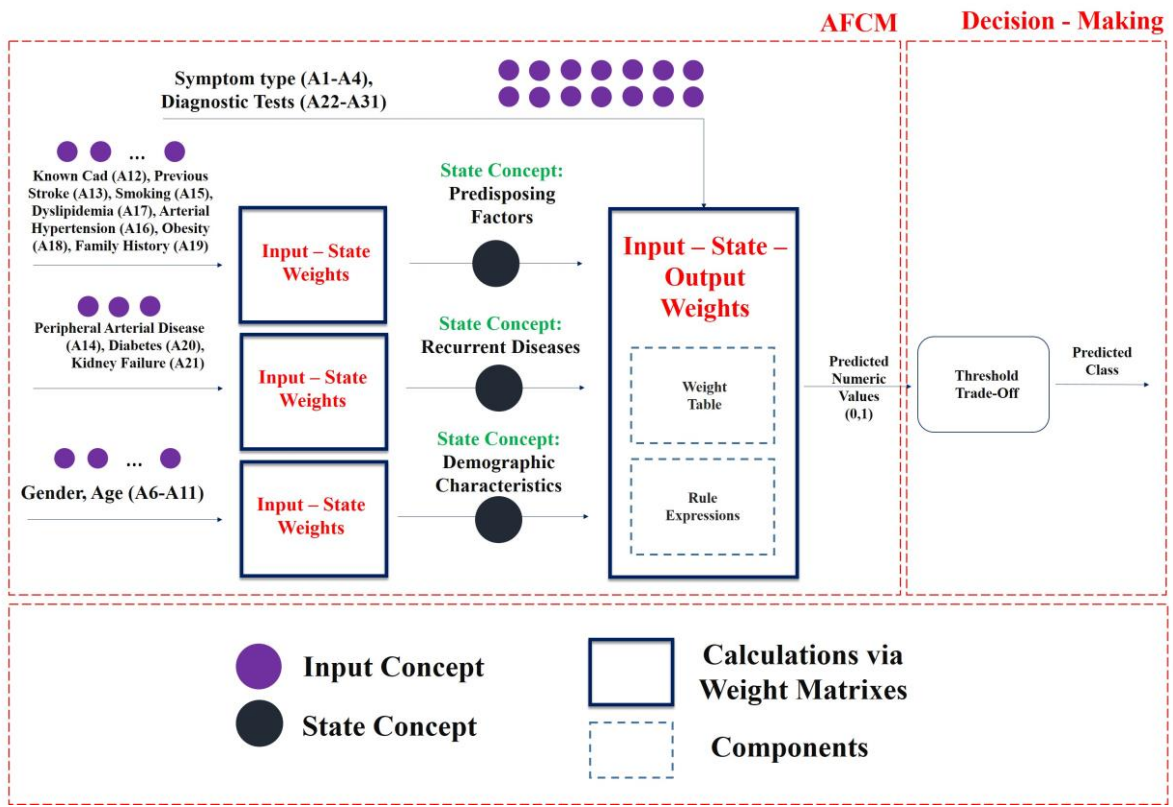

**Fig.4** Proposed Framework (RE-AFCM).

3.4. Comparisons with state-of-the-art algorithms

In this section we compare the results of the proposed system with state-of-the-art classification algorithms. The cross – validation was chosen to be 5-fold due to the small amount of training data. The Neural Network contains three depth layers of 1024, 512, and 256 nodes. Training was performed for 120 epochs, with a batch size of 32 and an initial learning rate of 0.01. Other Neural Networks with alterations on the pre-mentioned parameters were tested and excluded due to inefficiency. Hence, the Neural Network we pick is the one with the better accuracy over all. The Support Vector Machine (SMO) has the following parameters: Complexity Parameter (c) is set at 1.0, the value of epsilon is set at 1E-12, the batch size is 32 and the calibrator method is tuned to Logistic. AdaBoostM1 meta classifier is trained for 50 iterations, with a batch size of 32 and the decision classifier Decision Stump is selected. Chirp is trained with a batch size of 32. Random Forest was trained for 100 iterations with a bag size of 20%, and batch size 32. Spaarc was trained with a batch size of 32. The rest of the classifiers were trained with the optimal parameters suggested by MATLAB Machine Learning Toolkit.

**Table 9**. Machine Learning classifiers' results.

| Classifier | Accuracy (5-fold – cross validation) |
|---|---|
| Coarse Tree | 63.4 |
| Linear Discriminant | 70.3 |
| Logistic Regression | 70.6 |
| Linear SVM | 70.0 |
| Cubic SVM | 68.0 |
| Medium Gaussian | 73.3 |
| Coarse Gaussian | 62.7 |
| Medium KNN | 67.7 |
| Cubic KNN | 66.3 |
| Ensemble Bagged Trees | 68.3 |
| Ensemble Subspace Discriminant | 71 |
| Neural Network | 72.6 |
| Support Vector Machine (SMO) | 72.93 |
| AdaBoostM1 | 74.58 |
| Chirp | 76.89 |
| Spaarc | 72.93 |
| Random Forest | 74.58 |
| **RE-AFCM (this work)** | **85.47%** |

The results demonstrate that the proposed model, outperforms every Machine Learning classifier which was employed for the classification task in this work. The shortage of large-scale dataset is impeding the traditional machine learning algorithms to learn the actual and significant relationships between the concepts. This is reflected in the accuracy each algorithm obtains.

**4. Discussion**

It is demonstrated that the new approaches for the development of Advanced Fuzzy Cognitive Maps, as well as the rules mechanism inserted to the system, improve the results, compared to classic FCMs.

Fuzzy Cognitive Maps provide a distinguishable way to express the cause – effect relationship between phenomena, between numeric, nominal, binary, or categorical parameters. A complex system may include all the above-mentioned factors. Relationships between them, provided that are discovered, can be represented visually and mathematically through FCMs. In this way, not only the developers, but also the experts on the field, may observe and understand the FCM representations.

The interconnections between concepts learned by supervised learning algorithms are not ensured to reflect causal connections. That does not mean that unobserved causes may not exist; in fact, one target of supervised learning is to make assumptions regarding the relations between mutually affected concepts during the process of training and testing. Assumptions that can be confirmed or denied experimentally later. Compared to trainable artificial intelligence algorithms, AFCM do not intend to discover associations that may or may not exist.

The scientific community has enormously contributed towards the advance of FCMs, proposing a variety of FCM-based frameworks for the same or relative tasks. Some of those works are presented in Table 10.

**Table 10.** FCM-based methodologies

| Author / Publication | FCM-based proposal |
|---|---|
| Nair 2020 | Generalised Fuzzy Cognitive Maps (GFCMs) |
| Carvalho 1999 | Rule Based Cognitive Maps (RBFCM) |
| Belogianni 2018 | State-Space FCM |
| Wang 2020 | Deep FCM |
| Wu 2020 | Wavelet FCM |
| Papageorgiou 2019 | Hybrid FCM with Artificial Neural Network |
| Christoforou 2017 | Multi-layer FCM |

Future study opportunities include comparison of the various FCM-based approaches to a complete and pre-defined large-scale CAD dataset, in order to perform an in-depth analysis of the proposed systems.

Clinical researchers today are confronted with increasingly large, complex, and high-dimensional datasets (**Holzinger 2014**). Consequently, the application of interactive visual data exploration in combination with machine-learning techniques for knowledge discovery and data mining is indispensable.

Clinical researchers, or domain-experts are often not computer experts as well. They have high level medical domain-expert knowledge to perform their research, to interpret newly gained knowledge and patterns in their data. On the other hand, computer engineers lack the knowledge of medical data and their unique nature; Moreover, deep, dynamic and complex medical situations require a high level of expertise and experience for the decision making. A smooth interaction of the domain-expert with the data would greatly enhance the whole knowledge-discovery process chain (**Holzinger 2014**). This can be achieved by the cooperation of engineers and doctors.

That is the case in our work and, generally, in most of proposed models using AFCMs. The experts in a specific domain, not only have the overall supervision of the procedure, but also design the models in cooperation with the developers. Experts define the relationships, the concepts, the system's desired outputs. In our work, the doctor is in the loop, playing the most vital role in the development and the evaluation of the model.

Finally, the proposed model is explainable, as the user/expert can be notified about the degree of significance of each input, based upon which, the model yielded a specific result-prediction. Besides, interpretable and explainable algorithms are mandatory for medical decision-making systems (**Holzinger 2018**).

The State Space AFCM methodology needs further development from theoretical point of view: a) better separation of concepts, utilization of more experts, understand better the nonlinear behavior of the system, use learning methods to update the AFCM model, controllability and observability of the dynamic system, sensitivity of all concepts, b) the AFCM could be utilized to study other medical problems, especially the pandemic of coronavirus, c) develop appropriate software tools for the AFCM models, e) perform extensive simulation studies for as many as possible medical problems and validate these models, f) Develop Wise Learning (WL) methods and compare them with Deep Learning (DL) approaches of Artificial Intelligence (AI), g) employ the AFCM models, and feedback methodologies to develop Intelligent Control and Cognitive Control algorithms, and h) explore synergies between Neuroscience and fuzzy cognition.

## 5. Conclusion

In this research, a state approach of FCMs, as well as the cooperation of FCMs and rule-based decision mechanism, were applied and examined for the prediction of CAD. The proposed model was evaluated on a dataset of Coronary Artery Disease (CAD) candidates and outmatched several Machine Learning Algorithms regarding the prediction of CAD infection. The new approach combines the classical state space approach of dynamic systems to help improve the existing method of FCMs. The results of this study show that the new state space AFCM achieves better performance over the traditional FCM methodology. Therefore, we can conclude that the AFCMs' decision-making mechanism is a unique and promising method of handling the parameters of a difficult medical problem. To the best of our knowledge, seldom has State Space AFCM been applied to medical problems before (**Anninou 2017, 2018**). The rule-based implementation to the specific AFCM is an extension provided in the present research.

**Declarations**

Funding
The research received no external funding.

Conflicts of interest/Competing interests
The authors declare that there are no conflicts of interest.

Availability of data and material
The dataset is recorded at the Department of Nuclear Medicine of the University Hospital of Patras, Greece and is not allowed to be publically available.